\documentclass[letterpaper, 10 pt, journal, twoside]{template/IEEEtran}

\usepackage{graphicx}
\usepackage{caption}
\usepackage{amsmath,amssymb,enumerate}

\usepackage{amsthm}
\usepackage{mathtools}
\usepackage{breqn}
\usepackage{algorithm, algpseudocode}

\usepackage{blindtext}
\usepackage{gensymb}
\usepackage{xparse}
\usepackage{lipsum}
\usepackage{mathrsfs}
\usepackage{bm}
\usepackage[mathscr]{euscript}
\usepackage{times}
\usepackage[noadjust]{cite} 
\usepackage{multicol}
\usepackage{amsfonts}
\usepackage[utf8]{inputenc}
\usepackage[T1]{fontenc}
\usepackage{textcomp}
\usepackage{balance}
\usepackage{amsfonts}
\usepackage{soul}
\usepackage{multirow}
\usepackage{booktabs}
\usepackage{sidecap}
\usepackage{makecell}
\usepackage{array,float}

\usepackage{subcaption}
\usepackage[usenames,dvipsnames,svgnames,table, xcdraw]{xcolor}
\usepackage[colorlinks=true,pdfpagemode=UseNone,citecolor=black,linkcolor=black,urlcolor=BrickRed]{hyperref}

\newtheorem{remark}{Remark}

\renewcommand{\thefigure}{\arabic{figure}}

\definecolor{free}{rgb}{0.9608, 0.5882, 0.3922}
\definecolor{building}{rgb}{0.9608, 0.9020, 0.3922}
\definecolor{barrier}{rgb}{0.5882, 0.2353, 0.1176}
\definecolor{other}{rgb}{0.7059, 0.1176, 0.3137}
\definecolor{pedestrian}{rgb}{1, 0.3137, 0.3922}
\definecolor{pole}{rgb}{0.1176, 0.1176, 1}
\definecolor{road}{rgb}{0.7843, 0.1569, 1}
\definecolor{ground}{rgb}{0.3529, 0.1176, 0.5882}
\definecolor{sidewalk}{rgb}{1, 0, 1}
\definecolor{vegetation}{rgb}{1, 0.5882, 1}
\definecolor{vehicles}{rgb}{0.2941, 0, 0.2941}

\newcommand{\transpose}{\mathsf{T}}

\graphicspath{ {./media/} }

\begin{document}

\title{LatentBKI: Open-Dictionary Continuous Mapping in Visual-Language Latent Spaces with~Quantifiable~Uncertainty}

\author{Joey Wilson$^*$, Ruihan Xu$^*$, Yile Sun, Parker Ewen, Minghan Zhu, Kira Barton, and Maani Ghaffari%

\thanks{Manuscript received: October 3, 2024; Revised December 23, 2024; Accepted January 13, 2025.}
\thanks{This paper was recommended for publication by Editor Aleksandra Faust upon evaluation of the Associate Editor and Reviewers' comments.
This work was partly supported by the DARPA TIAMAT project.} 
\thanks{$^*$The authors contributed equally.}
\thanks{
The authors are with the University of Michigan, Ann Arbor, MI 48109, USA. {\texttt{\{wilsoniv,rhxu,sunyyyl,pewen\}@umich.edu}}
{\texttt{\{minghanz,bartonkl,maanigj\}@umich.edu}}
\text{(Corresponding Author: J. Wilson)}
}
\thanks{Digital Object Identifier (DOI): see top of this page.}
}

\markboth{IEEE Robotics and Automation Letters. Preprint Version. Accepted January, 2025}
{Wilson \MakeLowercase{\textit{et al.}}: LatentBKI: Open-Dictionary Continuous Mapping}

\maketitle



\begin{abstract}
This paper introduces a novel probabilistic mapping algorithm, LatentBKI, which enables open-vocabulary mapping with quantifiable uncertainty. 
Traditionally, semantic mapping algorithms focus on a fixed set of semantic categories which limits their applicability for complex robotic tasks.
Vision-Language (VL) models have recently emerged as a technique to jointly model language and visual features in a latent space, enabling semantic recognition beyond a predefined, fixed set of semantic classes.
LatentBKI recurrently incorporates neural embeddings from VL models into a voxel map with quantifiable uncertainty, leveraging the spatial correlations of nearby observations through Bayesian Kernel Inference (BKI). 
LatentBKI is evaluated against similar explicit semantic mapping and VL mapping frameworks on the popular Matterport3D and Semantic KITTI data sets, demonstrating that LatentBKI maintains the probabilistic benefits of continuous mapping with the additional benefit of open-dictionary queries. Real-world experiments demonstrate applicability to challenging indoor environments.
\end{abstract}


\IEEEpeerreviewmaketitle

\section{Introduction}

\IEEEPARstart{R}{obots} require informative world models to autonomously navigate the world, commonly known as maps. 
Mapping methods represent the geometry of the robot's surroundings and often include semantic information relevant to robotic task success.
While some works have proposed mapless autonomous navigation \cite{MaplessMP3, MaplessRL}, maps are commonly used in robotics due to the ability to leverage temporal information {\color{black}within an interpretable world model.}

Maps are also capable of storing a high level of scene understanding with multi-modal information such as occupancy, semantics, traversability, and uncertainty. 
As deep neural networks have rapidly progressed, mapping algorithms have also evolved from binary occupancy grids \cite{OccupancyGrid} to model higher levels of scene understanding such as through semantic information \cite{ConvBKI}.
However, real-world environments contain complex and detailed scenes that cannot be captured through closed-dictionary semantic maps.

\begin{figure}[t]
    \centering	
    \includegraphics[width=1\linewidth]{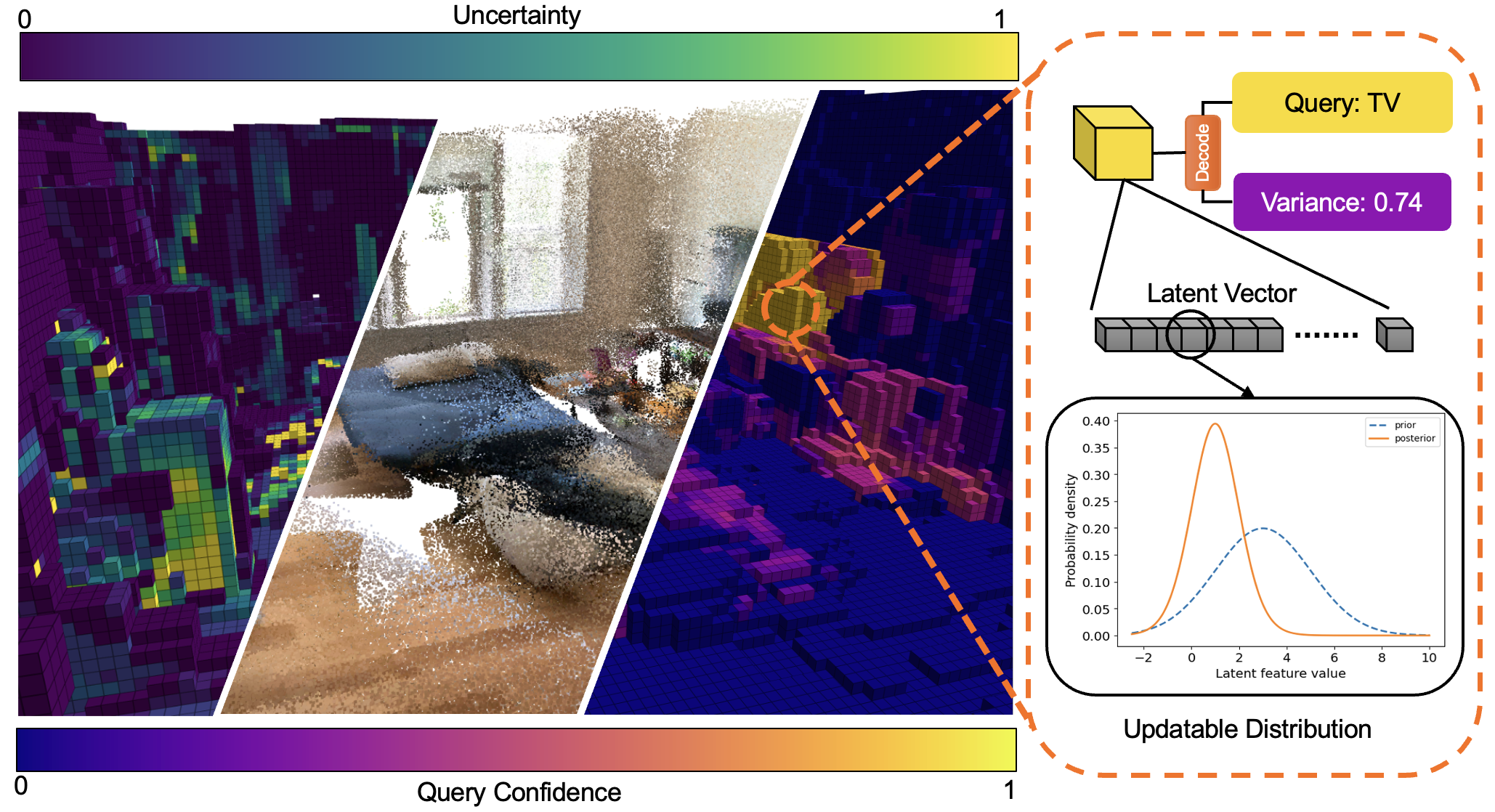}
	\caption{LatentBKI enables semantic mapping by leveraging open-dictionary language inference with vision-language (VL) model data. VL networks process exteroceptive data to generate point-wise features, which LatentBKI integrates into a 3D map via Bayesian Kernel Inference (BKI). Unlike prior VL mapping methods, LatentBKI updates nearby voxels using spatial information and maintains quantifiable uncertainty via conjugate priors. As shown, LatentBKI applies to real-world scenes (middle), quantifies semantic uncertainty per voxel (left), and decodes voxel features into categories using the VL network's language-driven decoder (right).}
    \label{fig:header}
    \vspace{-4mm}
    
\end{figure}

Recently, deep learning has produced foundation models trained on large, varied data sets with the reported ability to generalize to out-of-distribution data, solving a limitation of previous semantic segmentation neural networks \cite{kirillov2023segment, pmlr-v139-radford21a}. Of particular interest to the robotics community, Vision-Language (VL) models \cite{alayrac2022flamingo} extend vision processing to share a feature space with large language models, enabling applications such as open-dictionary semantic segmentation networks \cite{li2022languagedrivensemanticsegmentation}. {\color{black}Following the success of open-dictionary segmentation networks, several robotic mapping papers have proposed to integrate open-dictionary segmentation features within a robotic map in order to enable language-based queries and navigation \cite{gadre2022cowspasturebaselinesbenchmarks, huangvisuallanguagemapsrobot, AVLMap, chen2022openvocabularyqueryablescenerepresentations, jatavallabhula2023conceptfusionopensetmultimodal3d}. Within robotic open-dictionary mapping networks, a common approach is to fuse input points into the map through a simple moving average scheme, which does not require additional training \cite{huangvisuallanguagemapsrobot, AVLMap, jatavallabhula2023conceptfusionopensetmultimodal3d}. Inspired by the success of the simple averaging technique, we propose to  leverage the structure of continuous mapping to enable spatial smoothing and uncertainty quantification.}


{\color{black} Continuous mapping is an approach to probabilistic mapping which leverages Bayesian inference and spatial context to create complete maps with uncertainty \cite{GPOM, BKIOccupancy, MappingSBKI}. Quantifiable uncertainty is critical for robotics applications, as observations of the world are often limited and noisy, leading to errors and incomplete maps. Through uncertainty quantification in the form of variance, planning algorithms can identify potentially dangerous conditions \cite{ewen2024you} and the optimal trajectories to obtain new measurements \cite{ActivePerceptionNeural, ActivePerceptionNeural2}. Additionally, in the real world data is often sparse whether from sensors or views, leading to incomplete map representations. An efficient approach to continuous mapping known as Bayesian kernel inference (BKI) leverages spatial context to update nearby unobserved voxels through the use of an extended likelihood function defined by a kernel \cite{vega2014nonparametric}. BKI yields an efficient probabilistic update, however has not been studied as a possible solution for open-dictionary mapping.}



In this paper we propose an extension of continuous mapping to the latent space of neural networks called LatentBKI, which enables open-dictionary continuous mapping with spatial smoothing and quantifiable uncertainty. {\color{black}Compared to prior works in open-dictionary mapping, our method also leverages a weighted average to calculate the per-voxel expectation, however enables uncertainty quantification and fills in gaps within the map through BKI \cite{vega2014nonparametric}. Compared to prior research in continuous mapping for robotic maps, LatentBKI proposes a Gaussian likelihood conjugate pair for latent space measurements, enabling open-dictionary mapping and inference without any loss in performance. 

We evaluate our method against VLMap, which leverages a moving average recursive approach and is directly comparable to the expectation produced by LatentBKI.
We also quantitatively evaluate our method against closed-dictionary continuous mapping methods, demonstrating that LatentBKI enables open-dictionary mapping without losing performance and while maintaining the ability to quantify uncertainty.} 
Finally, we evaluate our method on real-world indoor scenes, highlighting the advantage of leveraging open-dictionary segmentation networks for continuous mapping. 
To summarize, our contributions are:

\begin{enumerate}
    \item Novel mapping algorithm which extends continuous Bayesian Kernel Inference (BKI) to latent spaces.
    \item Spatial smoothing and uncertainty quantification through conjugate priors in VL maps.
    \item Demonstration of segmentation and uncertainty quantification in real-world environments. 
    \item Open-source software is available for download at \href{https://github.com/UMich-CURLY/LatentBKI}{https://github.com/UMich-CURLY/LatentBKI}.
\end{enumerate}
\section{Related Work}

We review the literature on semantic mapping using continuous probabilistic inference, which creates comprehensive maps with quantifiable uncertainty but is limited to predefined categories. We then examine VL networks and maps, which allow for open-dictionary segmentation at inference time. LatentBKI addresses the challenge of integrating continuous probabilistic mapping with VL networks.

\subsection{Continuous Semantic Mapping}
Robots require advanced levels of scene understanding to plan, including knowledge of the geometry and semantic labels of objects and uncertainty associated with the objects to avoid failure due to mistaken object identity.
Often, these approaches use task-dependent object designations as semantic labels \cite{MappingSBKI} such as abstract topological information \cite{kuipers1991robot} or material classifications \cite{ewen2022these, ewen2024you}.
Robotics research has focused on incorporating segmentation predictions into maps via semantic label fusion \cite{mccormac2017semanticfusion, sunderhauf2017meaningful}.
Recent methods aim to quantify uncertainty through Bayesian inference \cite{ewen2022these, ewen2024you} by iteratively fusing semantic estimates projected onto a geometric map. {\color{black}Uncertainty quantification can be used by downstream planning algorithms to identify and circumvent potentially dangerous conditions \cite{ewen2024you}, as well as to plan optimally informative trajectories in a field of robotics known as active perception \cite{ActivePerceptionNeural, ActivePerceptionNeural2, placed2023surveyactivesimultaneouslocalization}.}

Kernel-based inference schemes have had notable success \cite{gan2022multitask, ConvBKI} in probabilistic semantic mapping.
Bayesian Kernel Inference (BKI), proposed by~\cite{vega2014nonparametric}, approximate the spatial influence of points at model selection through the usage of a kernel. BKI is an approximation of Gaussian Processes, which are effective at continuous mapping yet suffer from a cubic computational complexity \cite{GPOM, SemanticGP, OccupancyGP}.
Effectively, the kernel defines the shape or distribution of a point, deemed the extended likelihood, and can be applied to create an efficient, closed-form Bayesian update with more complete maps and quantifiable uncertainty \cite{BKIOccupancy}. 
While BKI has been applied effectively to semantic mapping \cite{MappingSBKI, ConvBKI, ConvBKIJournal}, semantic maps are inherently limited to a closed set of pre-specified categories. 
{\color{black}In contrast, we propose to extend the literature of continuous mapping to the latent space of neural networks through a Gaussian likelihood and conjugate pair, allowing for open-vocabulary inference within the latent space of VL models, without any loss in performance.}

\begin{figure*}[t]
    \centering
    \includegraphics[width=1\linewidth]{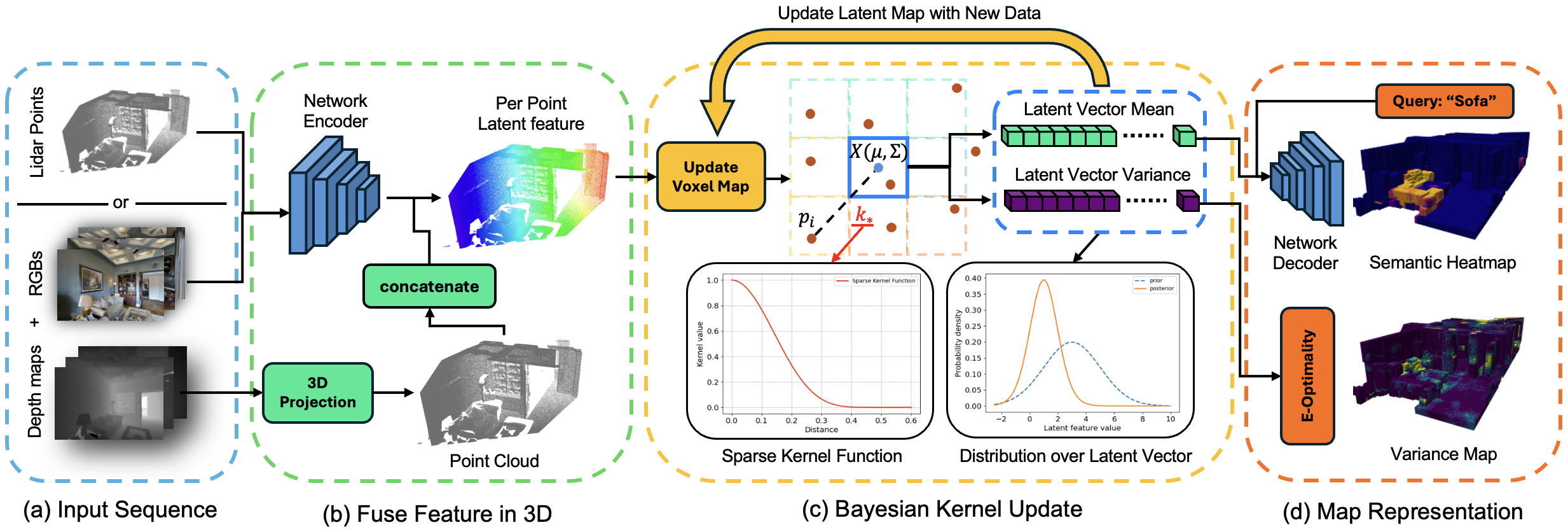}
    \caption{This figure demonstrates the overall pipeline of LatentBKI. (a) The input to LatentBKI is 3D points, which can be from LiDAR, RGB-D, or any exteroceptive sensor with 3D input. (b) Points are then processed by an off-the-shelf neural network, which encodes each point into a latent space. (c) By adopting a Gaussian likelihood over the point-wise features, we perform closed-form Bayesian inference on a voxel map where each voxel contains parameters modeling the conjugate prior of the multivariate Gaussian distribution. Additionally, instead of only considering points which fall within a voxel, we consider nearby points weighted through a kernel function. (d) The posterior predictive distribution of each voxel in latent space can then be decoded using the decoder of the neural network, enabling the computation of open-dictionary segmentation predictions with expectation and uncertainty.}
    \label{fig:latent-bki-pipeline}
    \vspace{-4mm}
\end{figure*}

\subsection{Vision Language Mapping}

Rapidly improving Large Language Models (LLMs) demonstrating remarkable generalizable capabilities have motivated the advent of vision-language models (VLM) with shared latent space for both images and texts \cite{alayrac2022flamingo, yao2022filip, li2022supervision}.
The pioneering method CLIP successfully represents visual and textual information in the same embedding dimension through contrastive learning
\cite{pmlr-v139-radford21a}.
Trained on a large dataset of image-text pairs, CLIP learns to embed features from visual or textual information in a shared feature space, where similarity is measured by a cosine similarity function
\cite{pmlr-v139-radford21a}. 

Based on the success of VLMs and their great zero-shot performance, recent robotics research has focused on open-dictionary mapping which operates in the latent space of VLMs and can create segmentation predictions from language descriptors \cite{li2022languagedrivensemanticsegmentation}. 
Approaches like LM-Nav
\cite{pmlr-v205-shah23b}, CoW \cite{gadre2022cowspasturebaselinesbenchmarks}, NLMap \cite{chen2022openvocabularyqueryablescenerepresentations} and VLMap \cite{huangvisuallanguagemapsrobot} have fused VLMs to enable robots to understand and navigate new environments. {\color{black}One common approach in literature to open-dictionary mapping is a volumetric averaging technique \cite{AVLMap, huangvisuallanguagemapsrobot, jatavallabhula2023conceptfusionopensetmultimodal3d} where images are processed through a language-driven semantic segmentation network such as LSeg \cite{li2022languagedrivensemanticsegmentation}, producing 3D points paired with neural features. Points are then incrementally fused within a volumetric map structure as a moving average of the features of points falling within each map cell. At inference time, the latent expectation of each voxel can then be decoded into per-category scores given the language embeddings of a set of categories, thereby enabling open-dictionary queries.
While successful, this approach loses the ability to quantify uncertainty and fill in gaps in the map from probabilistic continuous mapping, as discussed previously. 
Therefore, we propose to extend closed-dictionary continuous semantic mapping to open-dictionary VL maps to obtain quantifiable uncertainty and spatial smoothing.}

\section{Method}
We propose a novel method for probabilistic continuous mapping in the feature space of neural networks, which recurrently incorporates predictions from neural networks to learn an expectation and variance. Our mapping framework, which we call LatentBKI, has applications for general deep neural networks and is especially powerful when combined with modern foundation models such as VL models. 
Compared to previous methods which map in an explicit categorical space, continuous mapping in the feature space allows for open-dictionary queries with quantifiable uncertainty. 
A diagram of our method is shown in Fig. \ref{fig:latent-bki-pipeline}, demonstrating the ability of LatentBKI to complete scenes, decode semantic information, and quantify uncertainty in the latent space.

LatentBKI is built on the intuition that neural network features are geometrically continuous and suitable for kernel methods. Interpolation is a common step in modern neural networks to infer features from geometrically adjacent points, used especially in upsampling or deformable operations. Interpolation of point $x$ on feature grid $G$ with height $H$ and width $W$ can be written as:
\begin{equation}
    G_{x} \approx \frac{\sum_{i=1}^H \sum_{j=1}^W w_{ij} G_{ij}}{\sum_i \sum_j w_{ij}}
\end{equation}
\noindent where $i$ and $j$ are indices of neighboring cells, and weights $w$ are determined by the distance of query point $x$ to neighboring cells. This equation resembles the Nadaraya-Watson kernel estimate of the expected value, written similarly as:
\begin{equation}
    \hat{G_x} = \frac{\sum_{i=1}^N k(x - x_i) G_i}{\sum_{i=1}^N k(x - x_i)}
\end{equation}
\noindent for a set of $N$ data points. As we will show next, our method produces an expectation equivalent to the Nadaraya-Watson kernel estimate, with the addition of quantifiable uncertainty through conjugate priors. 

\subsection{Latent Mapping Representation} \label{sec:latent mapping representation}
Our map representation consists of a voxel map with voxels $*$ located at position {\color{black}$\bm{x}_*$} At each time-step our map is provided a set of points {\color{black}$\mathcal{D} = \{(\bm{x}_i, \bm{y}_i ) \}_{i=1}^N$}, where $x_i \in \mathbb{R}^{3}$ is the position of point $i$ and {\color{black}$\bm{y}_i \in \mathbb{R}^{C}$} is the corresponding latent feature of point $i$. From these points, our goal is to probabilistically update the latent parameters of each voxel, {\color{black}$\bm{y}_*$}, to obtain the \textit{posterior}.

In order to accomplish this goal, we first define a Gaussian likelihood over the feature space, such that: 
{\color{black}
    $p(\bm y_i | \bm \mu_i, \Sigma_i) = \mathcal{N}(\bm y_i ; \bm \mu_i, \Sigma_i)$. 
}
\noindent Since the features originate from a neural network, the likelihood defines an unknown expectation and variance which the point's features are sampled from. Similarly, we can define the points observed within a voxel $*$ according to the same likelihood distribution. From the likelihood, we can write an expression for the posterior of the latent parameters $\theta_* = \{\bm \mu_*, \Sigma_* \}$ of voxel $*$ using Bayes' rule as: {\color{black}
\begin{equation}\label{eq:Bayes Update}
    p(\theta_* \mid \bm x_*, \mathcal{D}) \propto p(\mathcal{D} \mid \theta_*, \bm x_*) p(\theta_* \mid \bm x_*).
\end{equation}
}
In order to model the distribution over the parameters $\theta_*$ of the voxel, which themselves define a multivariate Gaussian distribution, we adopt the conjugate prior of the multivariate Gaussian distribution, the normal-inverse Wishart distribution. The normal-inverse Wishart distribution defines a distribution over the multivariate Gaussian with unknown mean $\bm \mu_*$ and covariance $\Sigma_*$ through parameters $\bm \mu'_*, \Psi_*, \lambda_*, v_*$ as:
{\color{black}
\begin{equation}
    p(\bm \mu_*, \Sigma_*) = \mathcal{N}\left( \bm \mu_*; \bm \mu'_*, \frac{\Sigma_*}{\lambda_*} \right) \cdot \mathcal{W}^{-1}(\Sigma; \Psi_*, v_*),
\end{equation}
}
\noindent where the hyper-parameters represent the prior expectation of the mean {\color{black}($\bm \mu'_*$)}, the expectation of the covariance ($\Psi_*$), and the confidence in the mean and covariance estimates ($\lambda_*, v_*$). In our case, $\lambda_*$ and $v_*$ are equal and correspond to weighted counts of the total observations.

Although the conjugate prior provides a closed-form solution for updating the multivariate normal distribution parameters for each voxel, it does not consider the spatial locations of points. Intuitively, points that are closer to the centroid of the voxel should have a higher influence, while points that are further from the voxel centroid should have a lower influence. Additionally, only updating the voxels which points fall into can lead to sparse maps, as previously noted. Therefore, we adopt the solution of~\cite{vega2014nonparametric}, which defines an extended likelihood distribution that considers the spatial relationship of points to voxels through the use of a kernel function $k(\bm x_i, \bm x_*)$ as:
{\color{black}
\begin{equation}\label{eq:bki}
    p(\bm y_i | \bm x_i, \theta_*, \bm x_*) \propto p(\bm y_i | \theta_*)^{k(\bm x_i, \bm x_*)} .
\end{equation}
}
\noindent The only requirements when defining the extended likelihood are that $\color{black} k(\bm x, \bm x) = 1\ \forall x$ and $\color{black} k(\bm x, \bm x_*) \in [0, 1] \, \forall \, (\bm x, \bm x_*).$
Applied to the previously defined Gaussian likelihood, the extended likelihood can be written as: 
{\color{black}
\begin{equation}
    p(\bm y_i | \theta_*, \bm x_i, \bm x_*) = \mathcal{N} \left( \bm y_i ; \bm \mu_*, \frac{\Sigma_*}{k(\bm x_i, \bm x_*)} \right ). 
\end{equation}
}
{\color{black}\noindent Following Semantic BKI, we use a symmetric sparse kernel \cite{SparseKernel} with kernel length $l=0.5$ for direct comparison, where $d$ is the Euclidean distance between the two points:
\begin{align}
\label{eq:SparseKernel}
        \nonumber &k(d) = \\ 
        &\begin{cases}
        [\frac{1}{3} (2 + \cos (2 \pi \frac{d}{l})(1 - \frac{d}{l}) +
          \frac{1}{2 \pi} \sin (2 \pi \frac{d}{l})) ], 
          & \text{if } d < l \\
          0. & \text{else}
        \end{cases}
\end{align}}
Substituting the extended likelihood into \eqref{eq:Bayes Update}, we can now define a spatial update over the voxel parameters as:
{\color{black}
\begin{equation} \label{eq:bayes_extended}
    p(\theta_* | \bm x_*, \bm y_{1:N}, \bm x_{1:N}) \propto \left [ \prod_{i=1}^N p(\bm y_i | \theta_*, \bm x_i, \bm x_*) \right ] p(\theta_* | \bm x_*).
\end{equation}
}
Next, we present our map update algorithm, which follows the closed-form solution derived by~\cite{vega2014nonparametric}.

\subsection{Latent Mapping Update} \label{sec:latent mapping update}

First, we initialize the confidence over the mean and covariance of each voxel to a non-informative value of $\lambda_* \approx 0$. As points are observed, the value of $\lambda_*$ increases, indicating more confidence in the expected mean and covariance of the voxel. At time-step $t$, voxels are parameterized by prior $\bm \mu_*^{t-1}$, $\Psi_*^{t-1}$ and confidence $\lambda_*^{t-1}$, with input points $\mathcal{D}$. 

The influence of the new observations is calculated by:
{\color{black}
    $\Bar{k}_* = \sum_{i=1}^{N} k(\bm x_*, \bm x_i)$,
}
\noindent where the kernel function measures the influence of point $i$ over voxel $*$.
The confidence in the mean and covariance is updated:
    $\lambda_*^{t} = \lambda_*^{t-1} + \Bar{k}_*$.
Input observations are then used to compute the new mean as a running average:
{\color{black}
    $\Bar{\bm y}_* = \sum_{i=1}^{N} \frac{k(\bm x_*, \bm x_i)}{\Bar{k}_*}\bm y_i$, $ 
    \bm \mu_*^t = \frac{\lambda_*^{t-1} \bm \mu_*^{t-1} + \Bar{\bm y}_* \Bar{k}_*}{\lambda_*^{t}}$.
}

\begin{remark}
We note that the formulation of the new mean resembles the Nadaraya-Watson estimate mentioned in Section \ref{sec:latent mapping representation}, as input features are weighted by the kernel function to obtain a weighted average.
\end{remark}

Last, following \cite{vega2014nonparametric}, we update the expected covariance by weighting the covariance of the newly observed points:
{\color{black}
\begin{equation}
    \Bar{E}_* = (\Bar{\bm y}_* - \bm \mu_*^{t-1} )(\Bar{\bm y}_* - \bm \mu_*^{t-1})^T,
\end{equation}
\begin{equation}
    \Bar{S}_* = \sum_{i=1}^{N} k(\bm x_i, \bm x_*) (\bm y_i - \Bar{\bm y}_*)(\bm y_i - \Bar{\bm y}_*)^T,
\end{equation}
}
\begin{equation}
    \Psi*^{t} = \Psi*^{t-1} + \Bar{S}_* + \frac{\lambda_*^{t-1} \Bar{k}_*}{\lambda_*^t} \Bar{E}_*.
\end{equation}

\noindent As new points are obtained, the above process is repeated to update the mean, covariance, and confidence level. 



\subsection{Inference} \label{sectino:inference}
After updating the map, we can compute an expectation and variance for features observed within each voxel. First, the distribution can be marginalized to obtain a posterior predictive solution that defines the probability of observing a feature $\bm y_*$ at voxel centroid $\bm x_*$. The posterior predictive distribution for voxel $*$ is:
{\color{black}
\begin{equation}\label{eq:posterior predictive}
    p(\bm y_* | \bm x_*) = t_{\lambda_*} \left( \bm \mu_*, \frac{\lambda_* + 1}{\lambda_*^2}\Psi_* \right),
\end{equation}
}
\noindent where $t_{\lambda_*}$ is the multivariate Student-$t$ distribution. The multivariate Student-$t$ distribution has an expectation of:
    {\color{black}$\mathbb{E}(y_*) = \bm \mu_*$},
when $\lambda_* > 1$, and a covariance of: 
{\color{black}
    \mbox{$\text{Cov}(\bm y_*) = \frac{\lambda_*}{\lambda_* - 2} \left (\frac{\lambda_* + 1}{\lambda_*^2} \Psi_* \right )$},}
for $\lambda_* > 2$. However, both the expectation and covariance are within the latent space of the neural network and must be decoded to obtain a meaningful interpretation. 

When performing open-dictionary inference, the input is a set of phrases $W$ defining semantic categories, which are processed by a language model to obtain text embeddings $F_w \in \mathbb{R}^{C}$ per phrase $w \in W$. Specifically, in our experiment, we encode each phrase $w$ as a Clip feature vector $F_{w}$ of 512 length:
    $F_{w} = \text{Encoder}(w)$.
\noindent Inspired by LSeg, we obtain the categorical prediction $\hat{w}$ as: 
\begin{equation}
    \hat{w} = \mathrm{arg\,max}_{w}\, \frac{F_{w}^\transpose \mu_* }{\lVert \mu_*\rVert_{_2} \lVert F_{w}\rVert_{_2}} .
\end{equation}

\begin{remark}
We note that while we present the decoding for open-dictionary queries, LatentBKI can be decoded into any format using neural network decoders.
\end{remark}


\subsection{Uncertainty Quantification} \label{sec:uncertainty quantification}
While the map update step described above can propagate uncertainty, the covariance is limited to the latent space of the neural network encoder. Therefore, we propose two methods to quantify uncertainty.

First, following the approach of other neural network uncertainty quantification methods, we propose quantifying uncertainty through sampling. To quantify uncertainty through sampling, we sample many realizations of the voxel feature $y_*$, which we decode through the neural network decoder. Then, we compute the variance of the predictions in the decoded space. While this approach is accurate, it requires extra computation, which we propose to avoid.

{\color{black}Based on a common approach of information quantification in optimal experimental design \cite{placed2023surveyactivesimultaneouslocalization}, we propose to consolidate the covariance matrices as a single scalar value $U_*$ through p-optimality \cite{kiefer1974general}. Although p-optimality leads to many solutions \cite{placed2022general}, in this work we calculate and compare E-optimality, which selects the maximum eigen-value of the covariance matrix, and D-optimality, which calculates the volume of the covariance hyper-ellipsoid. For eigen-values $\bm \lambda$ of the covariance matrix, which we note are the diagonals of an uncorrelated covariance matrix, E-optimality criterion is computed as:
     $U_* = \max \left( \bm \lambda \right)$,
\noindent and D-optimality criterion is computed as: 
 \begin{equation}
     U_* = \exp \left(\frac{\sum_{i=1}^C \log \left( \bm \lambda_i \right)}{C} \right).
 \end{equation}
Experimentally, we find that E-optimality is highly correlated with the sampling-based uncertainty and is quick to compute. See Section~\ref{sec:variance exp} for detailed experiments}


{\color{black}
\subsection{Feature Compression}
Due to the large latent dimension of VLM's, we propose to make two approximations to reduce computation and memory complexity. First, we approximate the covariance matrix with only the diagonal elements, significantly reducing complexity at the cost of cross-correlation terms. Diagonal covariances are common in the feature space and are used in variational auto-encoders (VAEs). Second, we use PCA to reduce the latent dimension of encoded features from a latent dimension of 512 to 64 from VLM's before fusing into our map. PCA is an unsupervised learning algorithm that maximizes information preserved during compression and uses an affine transformation to upscale the compressed features back to the original dimension. Since the transformation is affine, we can compute the full dimensional expectation by passing the compressed expectation through the PCA upsampling. For the uncertainty, we compute E and D optimality in the reduced latent space and obtain full dimensional samples for the sampling technique through PCA upsampling. Overall, we found that PCA is generalizable to new scenes with minimal performance loss from compression.
}

\section{Results and Discussion}
We quantitatively and qualitatively show that LatentBKI effectively extends the continuous probabilistic mapping literature to neural network latent spaces, bringing quantifiable uncertainty and spatial smoothing to VL maps. First, we compare LatentBKI against closed-dictionary continuous mapping to verify that operations in the latent space of neural networks do not affect mapping performance. Next, we compare LatentBKI against VLMap, which performs latent space mapping but does not leverage spatial information or quantify uncertainty. Third, we study the correlation of our uncertainty quantification with segmentation errors and the effect of spatial smoothing. Finally, we conduct real-world experiments to demonstrate the ability of our map to transfer to real-world open-dictionary scenarios due to the strong generalization capabilities of large VL networks.

{\color{black}Quantitative results are obtained on popular outdoor and indoor datasets. For the outdoor results, we compare on the validation set of the Semantic KITTI dataset \cite{behley2019iccvkitti} using the Sparse Point-Voxel Convolution Neural Network (SPVCNN) \cite{tang2020searchingefficient3darchitectures} for point cloud semantic prediction. We choose the validation set because it has publicly available ground truth, and $4,070$ frames. For the indoor comparison, we evaluate methods on all eight scenes of the Matterport3D (MP3D) \cite{indoor-matterport3d} dataset  with an open-dictionary image semantic segmentation network, Language-driven Semantic Segmentation (LSeg) \cite{li2022languagedrivensemanticsegmentation}. Since the latent space of LSeg has a large dimension of 512, we use PCA to down-sample features for the map update to a size of 64.}



\begin{table}[b]
    \vspace{-4mm}
    \centering
    \caption{Comparison against closed-dictionary BKI mapping.}
    \begin{tabular}{c|c|c|c|c}
        Data & Method & Acc. $(\%)$ & mIoU $(\%)$ & Queries\\
        \hline \hline
        \multirow{3}{*}{Indoor} & Segmentation &  59.14 & 14.64 & N/A \\
        & ConvBKI (Single) & 61.49 & 16.69 & Fixed \\
        & LatentBKI & 60.44 & 16.15 & \textbf{Open} \\
        \hline \hline
        \multirow{3}{*}{Outdoor} & Segmentation &  89.60 & 58.54 & N/A \\
        & ConvBKI (Single) & 90.02 & 61.26 & Fixed \\
        & LatentBKI & 90.02 & 61.54 & \textbf{Open}
    \end{tabular}
    \label{tab:categorical_comp}
\end{table}

\subsection{Comparison against categorical space mapping method} \label{section: categorical comparison}

First, we compare LatentBKI against the closed-dictionary semantic mapping method defined in \cite{MappingSBKI}, which leverages BKI with a categorical likelihood to update the map. Specifically, we compare against ConvBKI \cite{ConvBKI} using a single un-trained spherical kernel for direct comparison. Our goal of this study is to verify that LatentBKI can generalize BKI into the latent space of neural networks without any significant changes in performance. {\color{black}Note that in this experiment, similar quantitative results indicate successful application of BKI to the latent space without any loss of functionality.}




We apply the same configuration to both mapping algorithms to ensure comparable results. Each algorithm uses a voxel resolution of 0.1 meters, a sparse kernel with a kernel length of 0.5 meters, and a filter size of 3, determining how many neighboring voxels should be updated for a single-point observation along a single axis. Since both methods compare spatial smoothing, we provide 80\% of the points as input and evaluate semantic predictions over the mean intersection over union (mIoU) and accuracy metrics on the remaining 20\% of the points.



As shown in Table \ref{tab:categorical_comp}, LatentBKI performs similarly to ConvBKI over indoor and outdoor datasets {\color{black} without any decrease in performance}. These results verify that {\color{black} our approach generalizes BKI to the latent space of networks, enabling open-dictionary probabilistic mapping, successfully}. While LatentBKI results in a marginal improvement in quantitative performance on the outdoor data, the slight decrease in indoor data is due to the dimensionality reduction applied by PCA on the input to LatentBKI. Next, we compare LatentBKI with a popular latent mapping algorithm, VLMap.

\subsection{Latent Mapping Comparison}
We compare LatentBKI with a similar open-dictionary mapping method VLMap \cite{huangvisuallanguagemapsrobot}, which also updates voxels through a weighted averaging approach. {\color{black}We choose to compare specifically with VLMap since weighted averaging is a popular technique for open-dictionary mapping \cite{jatavallabhula2023conceptfusionopensetmultimodal3d, AVLMap, huangvisuallanguagemapsrobot}, and the volumetric representation of VLMap allows for a direct and conclusive comparison. Since our approach generalizes VLMap to include a spatial kernel and quantifiable uncertainty, we compare the results with ($k=3$) and without ($k=1$) spatial smoothing, where $k$ is the number of neighboring voxels along each dimension an observation can influence. To demonstrate the benefits of VLMap, we also implement a heuristic baseline that stores the feature of the most recent coinciding observation within each voxel.} 

We compare each method on the MP3D dataset \cite{indoor-matterport3d} following the same experimental setup as VLMap, including a resolution of $0.05$ m to account for the fine-resolution indoor environment. Additionally, following the setup of VLMap we discard pixels with extreme depths $< 0.1$ m or $> 6$ m, discard points outside of the scene range, and downsample input points to the same set of $1\%$ of input pixels. By following the same downsampling heuristics as VLMap in our evaluation, we isolate the effect of the sparse kernel function and spatial smoothing used by our method to weight input points compared to the depth-wise weighting scheme employed by VLMap \cite{jatavallabhula2023conceptfusionopensetmultimodal3d, 6599048}.

{\color{black}The results of our experiments in Table \ref{tab:open_comp} indicate that both LatentBKI and VLMap outperform the heuristic baseline, demonstrating the benefit of the weighted average approach. LatentBKI outperforms VLMap in both accuracy and mean IoU, which we attribute to the sparse kernel and spatial smoothing. We note that our method is a probabilistic generalization of VLMap with a spatial kernel and quantifiable uncertainty, however these benefits also increase computational complexity linearly with the size of the spatial kernel. Although the computational complexity of our method is greater, our implementation is more efficient than VLMap due to vectorization and the use of the GPU, requiring $122.05$ ms to update $100,000$ points compared to $4,325.74$ ms for VLMap to update the same number of points.}



\begin{table}[t]
    \centering
    \caption{Latent mapping comparison on MP3D.}
    \begin{tabular}{c|c|c|c}
        Method & Acc. $(\%)$ & mIoU $(\%)$ & Uncertainty\\
        \hline \hline
        Segmentation \cite{li2022languagedrivensemanticsegmentation} & 53.24 & 12.59 & N/A \\
        Heuristic & 51.63 & 11.60 & No \\
        VLMap \cite{huangvisuallanguagemapsrobot} & 53.84 & 12.53 & No \\
        LatentBKI ($k=1$) & 55.57 & 14.01 & \textbf{Yes} \\
        LatentBKI ($k=3$) & \textbf{55.86} & \textbf{14.18} & \textbf{Yes} \\ 
        \hline \hline
    \end{tabular}
    \label{tab:open_comp}
    \vspace{-4mm}
\end{table}


\subsection{Ablation Studies}\label{sec:variance exp}

Two benefits of LatentBKI are the spatial smoothing effect of continuous mapping and the ability to quantify the temporal uncertainty of neural network predictions. In this section, we study the quantitative improvement from different kernel sizes, as well as compare the sampling and P-Optimality methods for quantifying uncertainty.

\begin{figure}[t]
    \centering
    \begin{subfigure}[b]{0.49\linewidth}
        \centering
        \includegraphics[width=\linewidth]{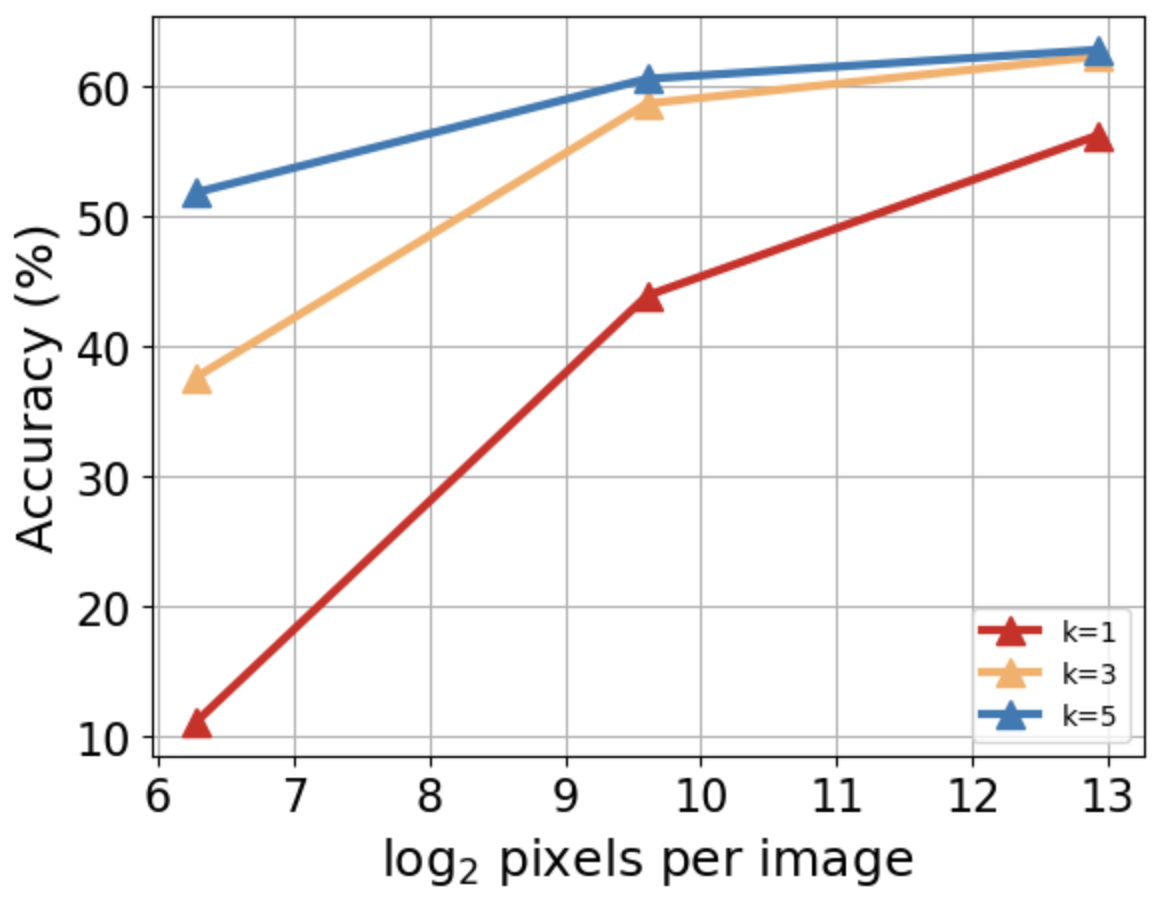}
        \caption{Accuracy vs. Input Sparsity}
    \end{subfigure}
    \begin{subfigure}[b]{0.49\linewidth}
        \centering
        \includegraphics[width=\linewidth]{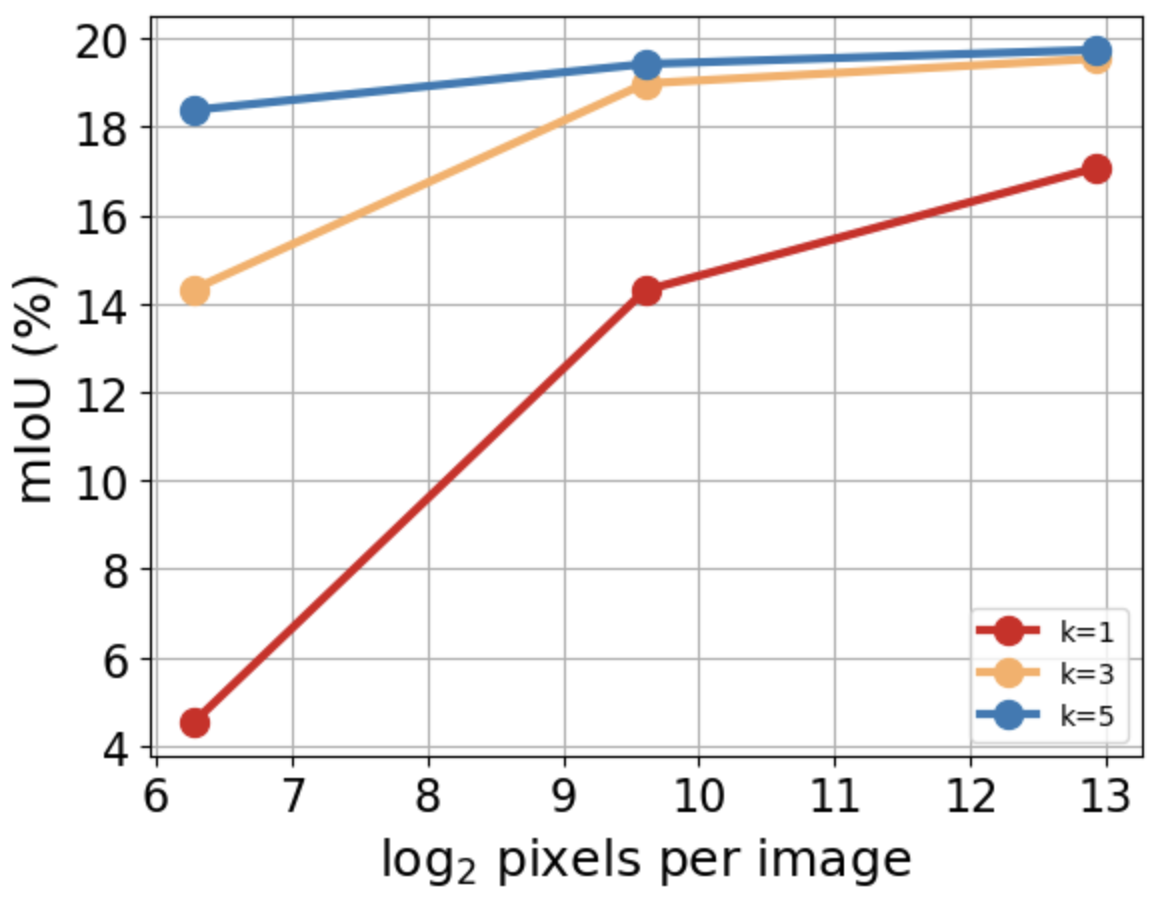}
        \caption{mIoU vs. Input Sparsity}
    \end{subfigure}
    \caption{Effect of spatial smoothing with varying levels of image sparsity. Spatial smoothing, indicated by the filter size $k$, is most effective for sparse images. The original image has a resolution of $720$ by $1080$ pixels.}
    \label{fig:performance vs sparsity}
    \vspace{-4mm}
\end{figure}

\begin{figure}[b]
    \centering
    \vspace{-4mm}
    \begin{subfigure}[b]{0.49\linewidth}
        \centering
        \includegraphics[width=\linewidth]{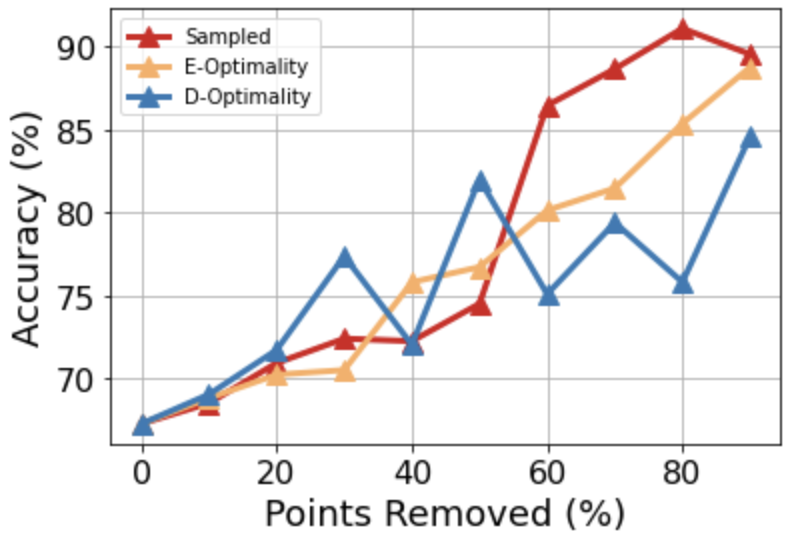}
        \caption{Accuracy vs. uncertainty}
    \end{subfigure}
    \begin{subfigure}[b]{0.49\linewidth}
        \centering
        \includegraphics[width=\linewidth]{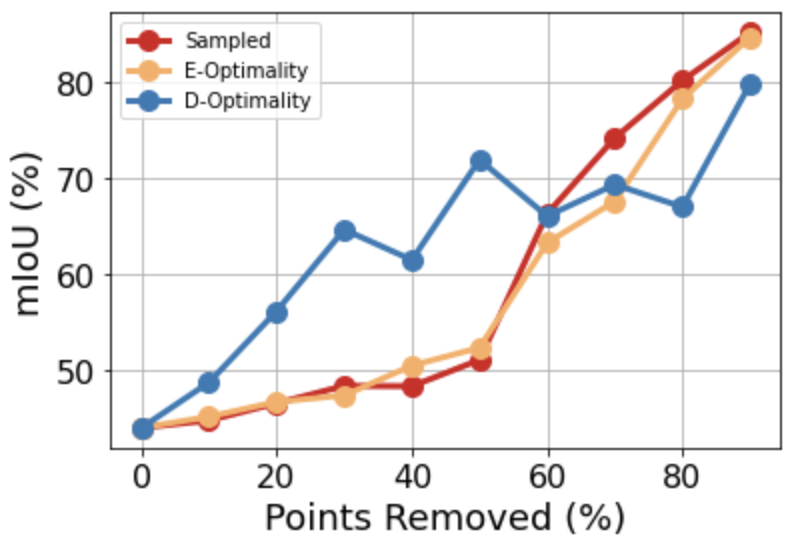}
        \caption{mIoU vs. uncertainty}
    \end{subfigure}
    \caption{Sparsification plot of segmentation performance compared to quantified uncertainty. As uncertain points are removed, a well calibrated uncertainty should cause the segmentation performance to increase.}
    \label{fig:uncertainty vs performance}
\end{figure}

\textbf{Spatial Smoothing:} In real-world applications, data is often sparse due to sensors such as LiDAR or sparse stereo matching algorithms. BKI provides a probabilistic technique to create more complete maps from sparse spatial data by leveraging the spatial smoothing effect of kernels. In this experiment, we compare different kernel sizes ($k$) and their ability to complete the map from sparse data.

All kernels are compared on the same scene of MP3D, where data is downsampled temporally to incorporate only one in 3 frames and at the image level to use a randomly sampled set of pixels from each image. Fig. \ref{fig:performance vs sparsity} demonstrates plots of the segmentation performance of different kernel sizes and varying sparsity levels. At extreme sparsity levels, spatial smoothing of points benefits the map completeness. As the input becomes more dense, the effect of the kernel is diminished.

\begin{figure}
    \centering
    \includegraphics[width=0.8\linewidth]{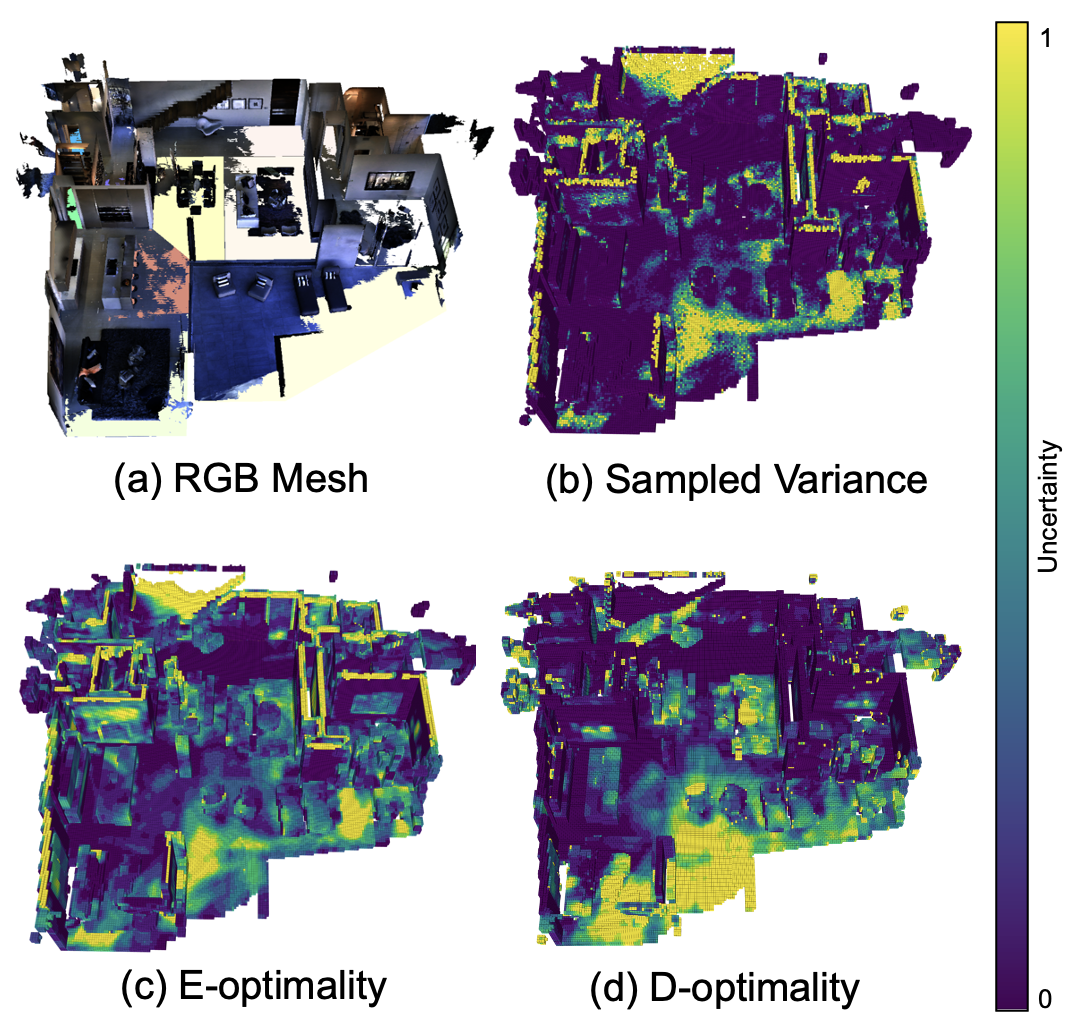}
    \caption{Uncertainty maps for 5LpN3gDmAk7
 MP3D sequence. (a) Covered house mesh in the sequence. (b) Categorical variance map by sampling from distribution. (c) Variance map by using E-optimality in latent space. (d) Variance map by using D-optimality in latent space.}
    \label{fig:uncertainty_map_visual}
    \vspace{-4mm}
\end{figure}

\textbf{Uncertainty Quantification:} To evaluate the ability of LatentBKI to quantify uncertainty meaningfully, we quantitatively and qualitatively compare uncertainty quantification on the MP3D dataset using LSeg as the encoder. We construct a map using LatentBKI, then compare uncertainty quantified using D-optimality, E-optimality, and sampling as described in Section \ref{sec:uncertainty quantification}. Whereas the sampling-based method is commonly used, it is computationally expensive compared to the E-optimality and D-optimality-based techniques {\color{black}with a run-time of $5,661$ ms for $10,000$ query voxels compared to $2.3$ ms for the optimality approaches to compute.}


To quantitatively compare each uncertainty quantification method, we create sparsification plots \cite{sparsification} identifying the correlation between uncertainty and prediction error, shown in Fig.~\ref{fig:uncertainty vs performance}. To create the sparsification plots, we sort points in a test set by the predicted uncertainty. Next, we separate the sorted points into bins and iteratively remove the most uncertain bin. If the uncertainty is properly calibrated, we expect to see an increase in the accuracy as uncertain points are removed. As seen in Fig.~\ref{fig:uncertainty vs performance}, both the accuracy and mIoU metrics are correlated with all three methods, especially sampling and E-optimality. In addition to a strong correlation between latent uncertainty and error in the decoded predictions, E-optimality benefits from efficient computation.

We also qualitatively compare uncertainty quantification between sampling and E-optimality, shown in Fig.~\ref{fig:uncertainty_map_visual}. We observe that the most uncertain voxels are typically located at the edges of rooms or at objects that are difficult for the VL network to identify due to ambiguity or poorly captured images. Similar to the sparsification plots, we find that E-optimality closely aligns with the uncertainty estimated through sampling, indicating that E-optimality is an effective approximation for the latent uncertainty.

\begin{figure}[b]
    \vspace{-4mm}
    \centering
    \includegraphics[width=1\linewidth]{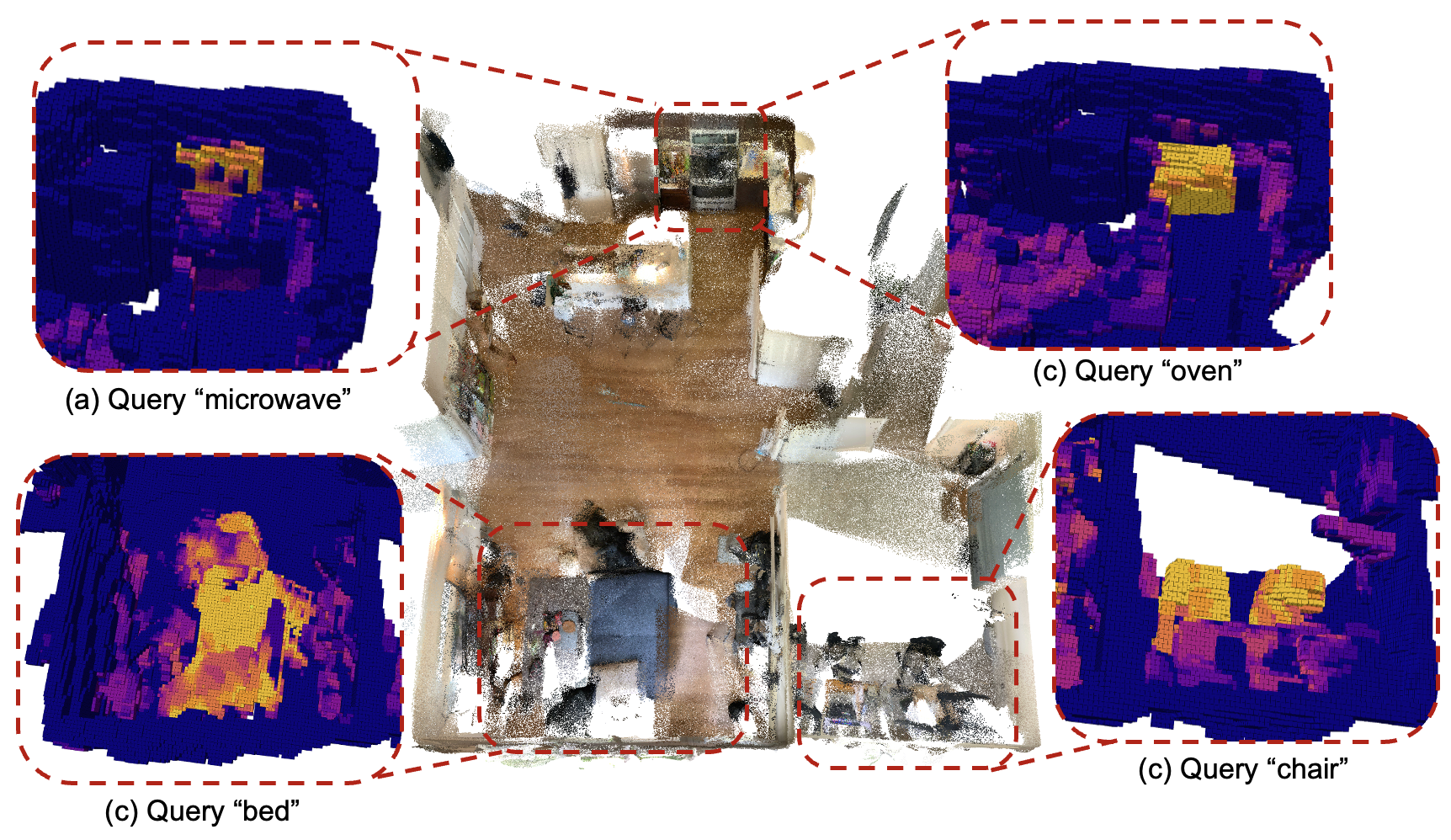}
    \caption{Open vocabulary query task results. Query results are shown in heat map, brighter colors indicates higher values.}
    \label{fig:open_vocabulary_application}
\end{figure}

\subsection{Real-World Experiment} \label{sec:real world exp}

To show that LatentBKI can generalize to real-world settings, we use an iPad with a 3D recording software, Record3D, to collect RGB-D data and camera poses of indoor scenes for mapping. We process the images with LSeg, and create a map of a real-world apartment using LatentBKI, shown in Figure \ref{fig:open_vocabulary_application}. 

In Fig. \ref{fig:open_vocabulary_application}, we demonstrate how Latent-BKI enables open-vocabulary queries which are more suitable for complex indoor environments. In this figure, we query arbitrary words in the map and portray a heat map of the voxels corresponding to the query word. While results were compared on a closed set of segmentation categories, our method enables language-based inference with quantifiable uncertainty. This is especially important because indoor environments can contain infinitely many categories of objects that cannot be captured adequately with a pre-specified set of objects.

 \begin{figure}[t]
    \centering
    \includegraphics[width=1\linewidth]{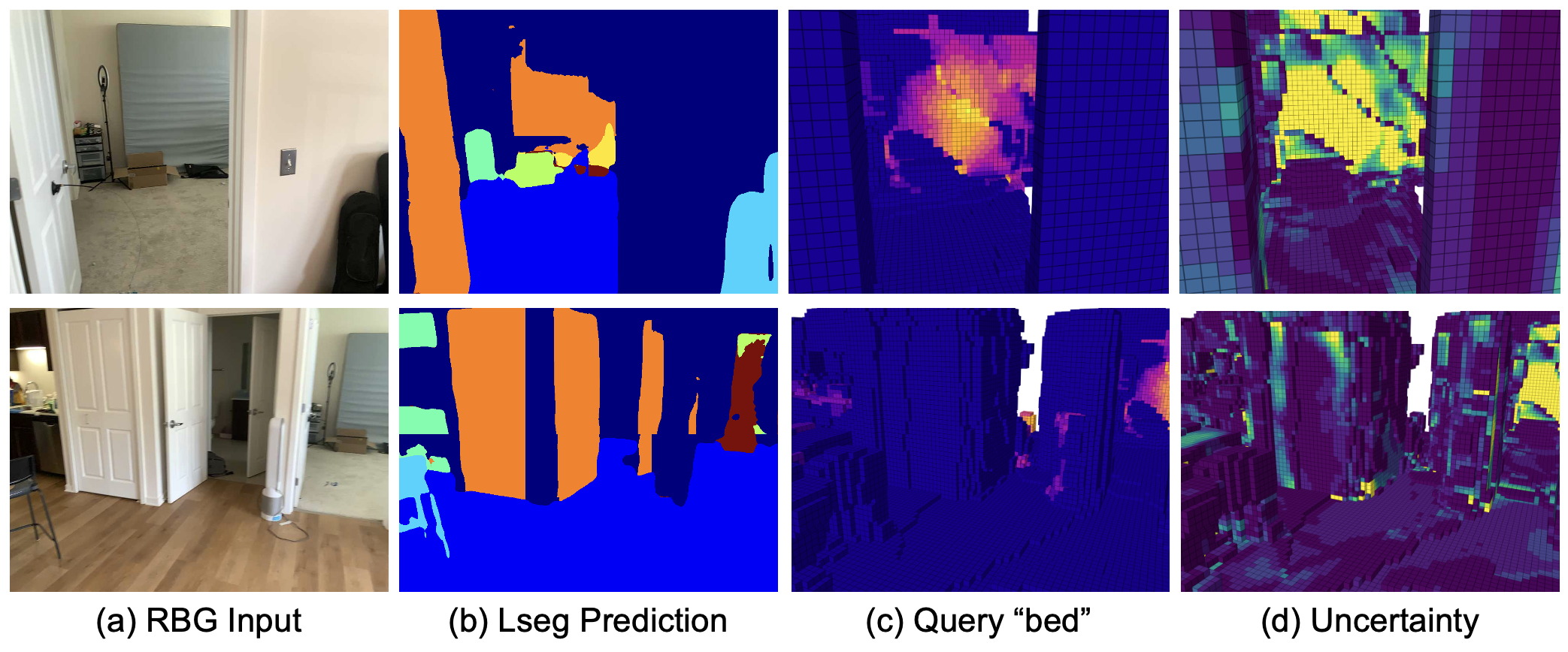}
    \caption{The first column (a) shows RGB input across different frames containing ``bed''. Column two (b) shows the Lseg network semantic prediction result, which gives inconsistent wrong semantic prediction across frames. Later two columns, (c) shows the heatmap of the query result of ``bed'' and (d) show the uncertainty. Our method shows a consistently high probability in the same area for the ``bed'' while maintaining the knowledge that the observations on the ``bed'' area are noise by showing high uncertainty.}
    \label{fig:inconsistent_bed}
    \vspace{-4mm}
\end{figure}

An additional benefit of LatentBKI is the ability to quantify uncertainty, which we demonstrate in Fig. \ref{fig:inconsistent_bed}. The input segmentation network has difficulty identifying a vertically placed mattress, producing inconsistent embeddings across different views. As a result, this region of the map exhibits high variance. Although the network prediction is noisy, LatentBKI can generate consistent query results for ``bed'' while acknowledging the high uncertainty from the network in that area.

\section{Conclusion}
\label{sec:conclusion}

We introduced LatentBKI, a novel method for probabilistically updating a voxel map where each voxel stores a latent descriptor in the embedding space of foundation models with quantifiable uncertainty. LatentBKI extends the classical literature of continuous semantic mapping to open-dictionary mapping, enabling language-based queries while maintaining quantifiable uncertainty. Language-based queries can handle the complexities posed by real-world robotic applications that may contain detailed environments and require human interaction. 

While LatentBKI demonstrated success in open-dictionary environments through a Gaussian likelihood, there are several avenues for future work. First, LatentBKI does not consider the unique geometry of objects and can therefore be combined with architectures such as ConvBKI, which learns per-category kernels, or the high-quality 3D Gaussian Splatting \cite{kerbl20233dgaussiansplattingrealtime} novel view synthesis method which represents the environment using 3D ellipsoids, similar to the kernel structure we leverage for continuous mapping. Additionally, inspired by the recent work on open-dictionary radiance fields \cite{qin2024langsplat3dlanguagegaussian}, we believe that the segment anything model \cite{kirillov2023segment} may be useful when identifying the boundaries of objects. {\color{black}Last, the uncertainty produced by LatentBKI may be used for planning tasks such as active perception by quantifying expected information gain \cite{ActivePerceptionNeural, ActivePerceptionNeural2, placed2023surveyactivesimultaneouslocalization}.}

\section*{ACKNOWLEDGMENT}

This work was partly supported by the DARPA TIAMAT project. M. Ghaffari thanks Dr. Alvaro Velasquez for the encouragement and support.


{\footnotesize
\balance
\bibliographystyle{template/IEEEtran}
\bibliography{bib/strings-abrv,bib/ieee-abrv,bib/refs}
}

\end{document}